# Convolutional Imputation of Matrix Networks

Qingyun Sun [*1]  Mengyuan Yan [*2]  David Donoho [3]  Stephen Boyd [2]

## Abstract

A matrix network is a family of matrices, with relatedness modeled by a weighted graph. We consider the task of completing a partially observed matrix network. We assume a novel sampling scheme where a fraction of matrices might be completely unobserved. How can we recover the entire matrix network from incomplete observations? This mathematical problem arises in many applications including medical imaging and social networks. To recover the matrix network, we propose a structural assumption that the matrices have a graph Fourier transform which is low-rank. We formulate a convex optimization problem and prove an exact recovery guarantee for the optimization problem. Furthermore, we numerically characterize the exact recovery regime for varying rank and sampling rate and discover a new phase transition phenomenon. Then we give an iterative imputation algorithm to efficiently solve the optimization problem and complete large scale matrix networks. We demonstrate the algorithm with a variety of applications such as MRI and Facebook user network.

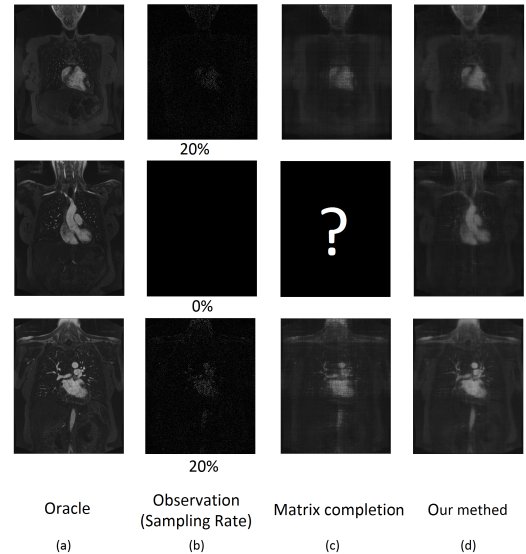

Figure 1: (a) Original MRI image frames (oracle). (b) Our sampled and corrupted observation. (c) Recovered image frames using nuclear norm minimization on individual frames. (d) Recovered image frames using our convolutional imputation algorithm.

## 1. Introduction

In machine learning and social network problems, information is often encoded in matrix form. User profiles in social networks can be embedded into feature matrices; item profiles in recommendation systems can also be modeled as matrices. Many medical imaging modalities, such as MRI and CT, also represent data as a stack of images. These matrices have underlying connections that can come from spatial or temporal proximity, or observed similarities between the items being described, etc. A weighted graph can be built to represent the connections between matrices. Therefore we propose matrix networks as a general model for data representation. A matrix network is defined by a weighted graph whose nodes are matrices.

[*]Equal contribution  [1]Department of Mathematics, Stanford University, California, USA  [2]Department of Electrical Engineering, Stanford University, California, USA  [3]Department of Statistics, Stanford University, California, USA. Correspondence to: Qingyun Sun <qysun@stanford.edu>.

Due to the limitations of data acquisition processes, sometimes we can only observe a subset of entries from each data matrix. The fraction of entries we observe may vary from matrix to matrix. In many real problems, a subset of matrices can be completely unobserved, leaving no information for ordinary matrix completion methods to recover the missing matrices. To our knowledge, we are the first to examine this novel sampling scheme.

As an example, in the following MRI image sequence (figure 1(a)), we sample each frame of the MRI images with i.i.d. Bernoulli distribution $p = 0.2$, and 2 out of 88 frames are completely unobserved, shown in figure 1(b). If we perform matrix completion by nuclear norm minimization on individual frames, we are not able to recover the completely unobserved matrices (figure 1(c)). When we build a network on the image frames, in this case, an one-dimensional chain representing the sequence, and assume that the matrices following graph Fourier transform are low-rank, we are able to recover the missing frames, as shown in figure 1(d).

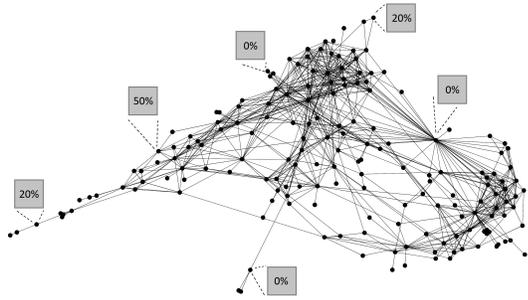

Figure 2: An example of matrix network on Facebook social graph. Each node on the graph represents a matrix.

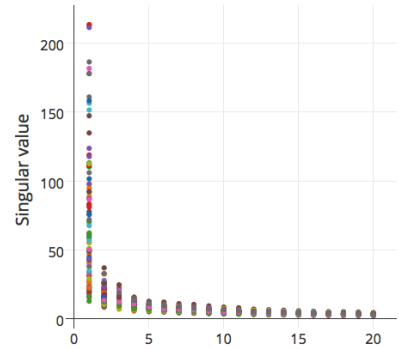

Figure 3: Spectral space singular value distribution of the MRI scan to support the spectral low-rank assumption.

The ability to recover all matrices from partial observations, especially inferring matrices that are totally unobserved, is crucial to many applications such as the cold start problem in networks. Illustrated in figure 2, new items or users in a network, which does not have much information available, need to aggregate information from the network to have an initial estimate of their feature matrices, in order to support inference and decisions.

Since we model the matrices as nodes on a graph, information from other matrices makes it possible to recover the missing ones. To use such information, we make the structural assumption that the matrix network is low-rank in spectral space, i.e., the matrix network is the graph convolution of two low-rank matrix networks. In the MRI example, we verify the spectral low-rank assumption in figure 3. For all the matrices after the graph Fourier transform, singular values quickly decrease to almost zero, demonstrating that they are in fact low-rank.

We make the following major contributions in this paper:

We define a novel modeling framework for collections of matrices using matrix network. We propose a new method to complete a stack of related matrices. We provide a mathematically solid exact recovery guarantee and numerically characterize the precise success regime. We give a convolutional imputation algorithm to efficiently complete large scale matrix networks.

## 2. Related work

Low-rank matrix recovery is an important field of research. (25; 26) proposed and (36; 48) improved the soft-impute algorithm as an iterative method to solve large-scale matrix completion problems. The soft-impute algorithm inspired our imputation algorithm. There was also a long line of works building theoretical tools to analyze the recovery guarantee for matrix completion (9; 7; 10; 8; 15; 22; 32; 31). Besides matrix completion, Gross analyzed the problem of efficiently recovering a low-rank matrix from a fraction of observations in any basis (24). These works enlightened our exact recovery analysis.

Low-rank matrix recovery could be viewed as a "non-commutative analog" of compressed sensing by replacing the sparse vector with a low-rank matrix. In compressed sensing, recovery of the sparse vector with a block diagonal sensing matrix was studied by the recent work (37), which demonstrated a phase transition that was different from the well-known phase transition for classical Gaussian/Fourier sensing matrices given by a series of works including (16; 3; 39). In our low-rank matrices recovery problem, our novel sampling scheme also corresponded to a block diagonal operator. We likewise demonstrated a new phase transition phenomenon.

Tensors could be considered as matrix networks when we ignored the network structure. Tensor completion coincides with the matrix network completion when the adjacent matrix is diagonal and the graph is just isolated points with no edge and the graph eigenbasis is the coordinate basis. Several works on tensor completion defined the nuclear norm for tensors as linear combinations of the nuclear norm of its unfoldings (21; 34; 19; 50; 49). Besides the common CP and Tucker decompositions of tensors, the recent work (51; 30; 35) defined the t-product using convolution operators between tensor fibers, which was close to the convolution of matrix network using the discrete Fourier transform matrix, and they applied the method in indoor localization. Departing from previous work, we considered a new sampling scheme in which some matrices were com-pletely unobserved, and the undersampling ratio could be highly unbalanced for the other observed matrices. This sampling scheme was not considered before, yet it is very natural under the matrix network model. Under an incoherent eigenbasis, we can recover one completely unobserved measurement matrix because its information is well spread across all the spectral matrices with low-rank structure.

Networks is an important modelling framework for relations and interactions (29; 20; 52; 54; 53). Graph Laplacian based regularization has been used in semi-supervised learning in (18; 1; 2; 38) and in PCA (44) and low-rank matrix recovery (41; 23). In (41; 23; 44) a regularization term is used for a single matrix, where both the row vectors and column vectors of the matrix are assumed to be connected with graphs.

The notion of graph Fourier transform is rooted in spectral graph theory (11), it is the cornerstone of graph harmonic analysis(13; 12). The coherence of graph Fourier transform is studied in (40; 46), and the examples of large graphs with low-coherence (non-local) eigenvectors include different classes of random graphs(14; 17; 47), and non-random regular graph(5). Graph Fourier transform and graph convolution are widely used in data analysis and machine learning, for example, in (4; 55; 28; 43; 45). Recent advances in the field of convolutional neural networks by (6; 27) used this idea to extend neural networks from working on Euclidean grids to working on graphs.

## 3. Mathematical definitions

**Matrix network.** First, consider a weighted graph $G$ with $N$ nodes and an adjacent matrix $W \in \mathbf{R}^{N \times N}$, where $W_{ij}$ is the weight on the edge between node $i$ and $j$. In the following, we use $J$ to represent the set of nodes on the graph.

We define a matrix network by augmenting this weighted graph $G$ with a matrix-valued function $A$ on the node set. The function $A$ maps each node $i$ in the graph to a matrix $A(i)$ of size $m \times n$. We define a $L_2$ norm $\|\cdot\|_2$ on the matrix network by the squared sum of all entries in all matrices of the network. And we define the sum of nuclear norm as $\|\cdot\|_{*,1}, \|A\|_{*,1} = \sum_{i=1}^{N} \|A(i)\|_*$.

**Graph Fourier transform.** The graph Fourier transform is an analog of the Discrete Fourier Transform. For a weighted undirected graph $G$ and its adjacent matrix $W$, the normalized graph Laplacian is defined as $L = I - D^{-1/2}WD^{-1/2}$, where $D$ is a diagonal matrix with entries $D_{ii} = \sum_j W_{ij}$. The graph Fourier transform matrix $U$ is defined using $UL = EU$, where $E$ is the diagonal matrix of the eigenvalues of $L$. Here, $U$ is a unitary $N \times N$ matrix, and the eigenvectors of $L$ are the row vectors of $U$. We rank the eigenvalues in descending order and identify the $k$-th eigenvalue with its index $k$ for simplicity.

For a matrix network $A$, we define its graph Fourier transform $\hat{A} = \mathcal{U}A$, as a stack of $N$ matrices in the spectral space of the graph. Each matrix is a linear combination of matrices on the graph, weighted by the graph Fourier basis. $\hat{A}(k) = \sum_{i \in J} U(k,i)A(i)$.

Intuitively, if we view the matrix network $A$ as a set of $m \times n$ scalar functions on the graph, the graph Fourier transform on matrix network is applying the graph Fourier transform on each function individually. Using tensor notation, the element of $A$ is $A(i, a, b)$, and the graph Fourier transform $\mathcal{U}$ can be represented by a big block diagonal matrix $U \otimes I$ where each block is $U$ of size $N^2$, and there are $(mn)^2$ such blocks.

We remark that the discrete Fourier transform is one special example of the graph Fourier transform. When the graph is a periodic grid, $L$ is the discrete Laplacian matrix, and the eigenvectors are just the basis vectors for the discrete Fourier transform, which are sine and cosine functions with different frequencies. We define the graph Fourier coherence as $\nu(U) = \max_{k,s} |U_{k,s}|$, following (40; 46). We know that $\nu(U) \in [\frac{1}{\sqrt{N}}, 1]$. When $\nu(U)$ is close to $\frac{1}{\sqrt{N}}$, the eigenvectors are non-local, for example, the discrete Fourier transform case, different classes of random graphs(14; 17; 47), and non-random regular graph(5). When $\mu(U)$ is close to 1, certain eigenvectors may be highly localized, especially when the graph has vertices whose degrees are significantly higher or lower than the average degree, say, in a star-like tree graph, or when the graph has many triangles, as discussed in (42). We will show in the following section that graphs with low coherence (close to $\frac{1}{\sqrt{N}}$) is preferred for the imputation problem.

**Convolution of matrix networks.** We can extend the definition of convolution to matrix networks. For two matrix networks $X, Y$ on the same graph, we define their convolution as
$$\widehat{(X \star Y)}(k) = \hat{X}(k)\hat{Y}(k).$$
Then $\widehat{X \star Y}$ is a stack of matrices where each matrix is the matrix multiplication of $\hat{X}(k)$ and $\hat{Y}(k)$. Convolution on a graph is defined as multiplication in the spectral space by generalizing the convolution theorem since it is not clear how to define convolution in the original space.

## 4. Completion problem with missing matrices

Imagine that we observe a few entries $\Omega(i)$ of each matrix $A(i)$. We define the sampling rates as $p_i = |\Omega(i)|/(mn)$. The projection operator $P_\Omega$ is defined to project the full matrix network to our partial observation by only retaining entries in the set $\Omega = \bigcup \Omega(i)$.

The sampling rate can vary from matrix to matrix. The main novel sampling scheme we include here is that a subset of matrices may be completely unobserved, namely $p_i = 0$. This sampling scheme almost has not been discussed in depth in the literature. The difficulty lies in the fact that if a matrix is fully unobserved, there is no information at all from itself for the recovery, therefore we must leverage the information from other observed matrices.

To focus on understanding the essence of this difficulty, it is worth considering the extreme sampling scheme where each matrix is either fully observed or fully missing, which we call node undersampling.

To recover missing entries, we need structural assumptions about the matrix network $A$. We propose the assumption that $A$ can be well-approximated by the convolution $X \star Y$ of two matrix networks $X, Y$ of size $m \times r$ and $r \times n$, for some $r$ much smaller than $m$ and $n$. We will show that under this assumption, accurate completion is possible even if a significant fraction of the matrices are completely unobserved.

We formulate the completion problem as follows. Let $A^0 = X^0 \star Y^0$ be a matrix network of size $m \times n$, as the ground truth, where $X^0$ and $Y^0$ are matrices of size $m \times r$ and $r \times n$ on the same network. After the graph Fourier transform, $\hat{A}^0(k)$ are rank $r$ matrices. Our observations are $A^\Omega = P_\Omega(A) = P_\Omega(A^0 + W)$, each entry of $W$ is sampled i.i.d from $N(0, \sigma^2/n)$.

We first consider the noiseless setting where $\sigma = 0$. We can consider the following nuclear norm minimization problem, as a convex relaxation of rank minimization problem,

$$\begin{aligned} \underset{\hat{M}}{\text{minimize}} \quad & \|\hat{M}\|_{*,1}, \\ \text{subject to} \quad & A^\Omega = P_\Omega(\mathcal{U}^* \hat{M}) \end{aligned}$$

As an extension to include noise, we can consider the convex optimization problem in Lagrange form with regularization parameters $\lambda_k$,

$$L_\lambda(\hat{M}) = \frac{1}{2}\|A^\Omega - P_\Omega \mathcal{U}^* \hat{M}\|_2^2 + \sum_{k=1}^N \lambda_k \|\hat{M}(k)\|_*.$$

We can also consider the bi-convex formulation, which is to minimize the following objective function,

$$\begin{aligned} L_\lambda(X, Y) &= \|A^\Omega - P_\Omega(X \star Y)\|_2^2 \\ &+ \sum_{k=1}^N \lambda_k(\|\hat{X}(k)\|_2^2 + \|\hat{Y}(k)\|_2^2). \end{aligned}$$

This formulation is non-convex but it is computationally efficient in large-scale applications.

One remark is that when we choose the regularization parameter $\lambda_k$ to be $E_k$, the eigenvalues of the graph Laplacian $L$, and view $X$ as a $(nr) \times N$ dimensional matrix,

$$\sum_{k=1}^N E_k \|\hat{X}(k)\|_2^2 = \text{Tr}(X^* U^* EUX) = \text{Tr}(X^* LX),$$

then our regularizer is related to the graph Laplacian regularizer from (41; 23; 44).

## 5. Exact recovery guarantee

Let us now analyze the theoretical problem: what condition is needed for the non-uniform sampling $\Omega$ and for the rank of $\hat{A}$ such that our algorithm is guaranteed to perform accurate recovery with high probability? We focus on the noiseless case, $\sigma = 0$, and for simplicity of results, we assume $m = n$.

We first prove that one sufficient condition is that **average sampling rate** $p = \frac{1}{N}\sum_{i=1}^N p_i = |\Omega|/(Nn^2)$ is greater than $O(\frac{r}{n}\log^2(nN))$. It is worth pointing out that the condition is only about the average sampling rate, therefore it includes the interesting case that a subset of matrices is completely unobserved.

**Analysis on exact recovery guarantee** The matrix incoherence condition is a standard assumption for low-rank matrix recovery problems.

Let the SVD of $\hat{A}(k)$ be $\hat{A}(k) = V_1(k)E(k)V_2^*(k)$. We define $P_1$ and $P_2$ as the direct sum of the projection matrix, $P_1(k) = V_1(k)V_1(k)^*$, $P_2(k) = V_2(k)V_2(k)^*$. We define the subspace $T$ as the direct sum of the subspaces $T(k)$, where $T(k)$ is spanned by the column vectors of $V_1(k)$ and $V_2(k)$. Then we define the projection onto $T$ as $P_T(\hat{M}) = (V_1 V_1^* \hat{M} + \hat{M} V_2 V_2^* - V_1 V_1^* \hat{M} V_2 V_2^*)$. We define its complement as $P_{T^\perp} = I - P_T$. We define $\text{sign}(\hat{A}(k)) = V_1(k)V_2^*(k)$ as the sign matrix of the singular values of $\hat{A}$.

In matrix completion, for a $r$-dimensional subspace of dimension $n$, spanned by $V$ with an orthogonal projection $P_V$, the coherence $\mu(V)$ is defined as

$$\mu(V) = \frac{n}{r} \max_i \|P_V e_i\|^2.$$

Here we introduce an averaged coherence

**Definition 5.1** *For the graph Fourier transform $U^*$, let the column vector of $U^*$ be $u_k$. We now define the incoherence condition with coherence $\mu$ for the stack of spectral subspaces $V_1(k), V_2(k)$ as*

$$\max\{\sum_{k=1}^N \|u_k\|_\infty^2 \mu(V_1(k)), \sum_{k=1}^N \|u_k\|_\infty^2 \mu(V_2(k))\} = \mu.$$

The coherence of graph Fourier transform $U^*$ is defined as

$\nu(U^*) = \max_k \|u_k\|_\infty$. We remark that

$$\begin{aligned} \mu &\leq \nu(U^*)^2 \max\{\sum_{k=1}^N \mu(V_1(k)), \sum_{k=1}^N \mu(V_2(k))\} \\ &\leq \nu(U^*)^2 N \max_{k=1}^N \max\{\mu(V_1(k)), \mu(V_2(k))\} \end{aligned}$$

In the following, we show that the sampling rate threshold is proportional to $\mu$, which is upper bounded by $\nu(U^*)^2 N \max_{k=1}^N \max\{\mu(V_1(k)), \mu(V_2(k))\}$. This upper bound suggests that for the imputation would prefer low coherence graph such that $\nu(U^*)$ is close to $\frac{1}{\sqrt{N}}$.

**Theorem 1** *We assume that $A$ is a matrix network on a graph $G$, and its graph Fourier transform $\hat{A}(k)$ are a sequence of matrices, each of them is at most rank $r$, and $\hat{A}$ satisfy the incoherence condition with coherence $\mu$. And we observe a matrix network $A^\Omega$ on the graph $G$, for a subset of node in $\Omega$ random sampled from the network, node $i$ on the network is sampled with probability $p_i$, we define the average sampling rate $p = \frac{1}{N}\sum_{i=1}^N p_i = |\Omega|/(Nn^2)$, and define $\mathcal{R} = \frac{1}{p}P_\Omega \mathcal{U}^*$.*

*Then we prove that for any sampling probability distribution $\{p_i\}$, as long as the average sampling rate $p > C\mu \frac{r}{n} \log^2(Nn)$ for some constants $C$, the solution to the optimization problem*

$$\begin{aligned} \underset{\hat{M}}{\text{minimize}} & \quad \|\hat{M}\|_{*,1}, \\ \text{subject to} & \quad A^\Omega = \mathcal{R}\hat{M} \end{aligned}$$

*is unique and is exactly $\hat{A}$ with probability $1 - (Nn)^{-\gamma}$, where $\gamma = \frac{\log(Nn)}{16}$.*

**Proof sketch** The proof of this theorem is given in the supplementary material. We sketch the steps of the proof here.

- We will prove that for any nonzero $\hat{M} \neq \hat{A}$, we define $\Delta = \hat{M} - \hat{A}$, then we want to show either $\mathcal{R}\Delta \neq 0$, or $\|\hat{A} + \Delta\|_{*,1} > \|\hat{A}\|_{*,1}$. We define the inner product: $\langle \hat{M}_1, \hat{M}_2 \rangle = \sum_k \langle \hat{M}_1(k), \hat{M}_2(k) \rangle$, then $\|\hat{A}\|_{*,1} = \langle \text{sign}(\hat{A}), \hat{A} \rangle$. We define a decomposition $\Delta = P_T \Delta + P_{T^\perp} \Delta \triangleq \Delta_T + \Delta_T^\perp$. For $\mathcal{R}\Delta = 0$, we show that

$$\|\hat{A} + \Delta\|_{*,1} \geq \|\hat{A}\|_{*,1} + \langle \text{sign}(\hat{A}) + \text{sign}(\Delta_T^\perp), \Delta \rangle.$$

- Now we want to estimate

$$s_\Delta \triangleq \langle \text{sign}(\hat{A}) + \text{sign}(\Delta_T^\perp), \Delta \rangle.$$

Since $\mathcal{R}\Delta = 0$, $\Delta \in \text{range}(\mathcal{R})^\perp$. We want to construct a dual certificate $K \in \text{range}(\mathcal{R})$, such that for $k = 3 + \frac{1}{2}\log_2(r) + \log_2(n) + \log_2(N)$, with probability $1 - (Nn)^{-\gamma}$,

$$\begin{aligned} \|P_T(K) - \text{sign}(\hat{A})\|_2 &\leq (\tfrac{1}{2})^k \sqrt{r}, \\ \|P_{T^\perp}(K)\| &\leq \tfrac{1}{2}. \end{aligned}$$

- Given the existence of the dual certificate, we have

$$s_\Delta = \langle \text{sign}(\hat{A}) + \text{sign}(\Delta_T^\perp) - K, \Delta \rangle$$

We can break down $s_\Delta$ as

$$\langle \text{sign}(\Delta_T^\perp) - P_{T^\perp}(K), \Delta_T^\perp \rangle + \langle \text{sign}(\hat{A}) - P_T(K), \Delta_T \rangle$$

then with probability $1 - (Nn)^{-\gamma}$, we get

$$s_\Delta \geq \|\Delta_T^\perp\|_{*,1} - \frac{1}{2}\|\Delta_T^\perp\|_2 - (\frac{1}{2})^k \sqrt{r}\|\Delta_T\|_2.$$

- We can show that for all $\Delta \in \text{range}(\mathcal{R})^\perp$, with probability $1 - (Nn)^{-\gamma}$,

$$\|\Delta_T\|_2 < 2nN\|\Delta_T^\perp\|_2.$$

Using this fact,

$$s_\Delta \geq \frac{1}{2}\|\Delta_T^\perp\|_2 - (\frac{1}{2})^k \sqrt{r} 2nN\|\Delta_T^\perp\| \geq \frac{1}{4}\|\Delta_T^\perp\|_2$$

Therfore, when $\hat{M}$ is a minimizer, we must have $\Delta_T^\perp = 0$, otherwise $\|\hat{A} + \Delta\|_{*,1} < \|\hat{A}\|_{*,1}$. Since $\|\Delta_T\|_2$ is bounded by $\|\Delta_T^\perp\|_2$, we also have $\Delta_T = 0$, then $\Delta = 0$. Therefore, $\hat{M}$ is the unique mininizer, and $\hat{M} = \hat{A}$. This ends the proof.

- Now we add remarks for some of the important techinical steps. The propositions with high probability guarantee rely on a concentration result. Since $E(P_T \mathcal{R} P_T) = P_T$, we control the probability of deviation $\mathbf{P}[\|P_T - P_T \mathcal{R} P_T\| > t]$ via operator-Bernstein inequality( see theorem 6 of (24)), use the condition $p = C\mu \frac{r}{n} \log^2(Nn)$, let $t = 1/2$, then with probability $1 - (nN)^{-\gamma}$, where $\gamma = \frac{\log(Nn)}{16}$, $\|P_T - P_T \mathcal{R} P_T\| < 1/2$.

- We construct a dual certificate via a method called "golfing", this technique was invented in (24). We construct the dual certificate $K$ by the following construction: We decompose $\Omega$ as the union of $k$ subset $\Omega_t$, where each entry is sampled independently so that $E(|\Omega_t|) = p_t = 1 - (1-p)^{1/k}$, and define $R_t = \frac{1}{p_t} P_{\Omega_t} \mathcal{U}^*$. Define $H_0 = \text{sign}(\hat{A})$, $K_t = \sum_{j=1}^t R_j H_{j-1}$, $H_t = \text{sign}(\hat{A}) - P_T K_t$. Then the dual certificate is defined as $K = K_k$. Using the operator-Bernstein concentration inequality, we can verify the two conditions:

The first condition: $\|P_T(K) - \text{sign}(\hat{A})\|_2 = \|H_k\| \leq \|P_T - P_T \mathcal{R} P_T\| \|H_{t-1}\|_2 \leq \frac{1}{2}\|H_{t-1}\|_2 \leq (\frac{1}{2})^k \|\text{sign}(\hat{A})\| \leq (\frac{1}{2})^k \sqrt{r}$.

The second condition, we can apply operator-Bernstein inequality again for a sequence of $t_j = 1/(4\sqrt{r})$, so that $\|P_{T^\perp} R_j H_{j-1}\| \leq t_i \|H_{j-1}\|_2$, and since $\|H_j\|_2 \leq \sqrt{r} 2^{-j}$, then $\|P_{T^\perp}(K)\| \leq \sum_{j=1}^k t_i \|H_{j-1}\|_2 \leq \frac{1}{4} \sum_{j=1}^k 2^{-(j-1)} < 1/2$.

## 6. Convolutional imputation algorithm

Now we propose a convolutional imputation algorithm that effectively finds the minimizer of the optimization problem for a sequence of regularization parameters.

**Iterative imputation algorithm.** The vanilla version of our imputation algorithm iteratively performs imputation of $A^{\text{impute}} = P_\Omega(A) + P_\Omega^\perp(A^{\text{est}})$ and singular value soft-threshold of $\hat{A}^{\text{impute}}$ to solve the nuclear norm regularization problem. In the following, we denote singular value soft-threshold as $S_\lambda(\hat{A}) = V_1(\Sigma - \lambda I)_+ V_2^*$, where $(\cdot)_+$ is the projection operator on the semi-definite cone, and $\hat{A} = V_1 \Sigma V_2^*$ is the singular value decomposition.

---

Iterative Imputation:
**input** $P_\Omega(A)$.
  Initialization $A_0^{\text{est}} = 0$, $t = 0$.
  **for** $\lambda^1 > \lambda^2 > \ldots > \lambda^C$, where $\lambda^j = (\lambda_k^j), k = 1, \ldots, N$ **do**
    **repeat**
      $A^{\text{impute}} = P_\Omega(A) + P_\Omega^\perp(A_t^{\text{est}})$.
      $\hat{A}^{\text{impute}} = \mathcal{U} A^{\text{impute}}$.
      $\hat{A}_{t+1}^{\text{est}}(k) = S_{\lambda_k^j}(\hat{A}^{\text{impute}}(k))$.
      $A_{t+1}^{\text{est}} = \mathcal{U}^{-1} \hat{A}_{t+1}^{\text{est}}$.
      t=t+1.
    **until** $\|A_t^{\text{est}} - A_{t-1}^{\text{est}}\|^2 / \|A_{t-1}^{\text{est}}\|^2 < \epsilon$.
    Assign $A_{\lambda^j} = A_t^{\text{est}}$.
  **end for**
**output** The sequence of solutions $A_{\lambda^1}, \ldots, A_{\lambda^C}$.

---

In the vanilla imputation algorithm, computing the full SVD on each iteration is very expensive for large matrices. For efficiency, we can use alternating ridge regression to compute reduced-rank SVD instead. Due to the limited space, we omit the detailed algorithm description here.

**Regularization path.** The sequence of regularization parameters is chosen such that $\lambda_k^1 > \lambda_k^2 > \ldots > \lambda_k^C$ for each $k$. The solution for each iteration with $\lambda^s$ is a warm start for the next iteration with $\lambda^{s+1}$. Our recommended choice is to choose $\lambda_k^1$ as the largest singular value for $\hat{A}^{\text{impute}}(k)$, and decay $\lambda^s$ at a constant speed $\lambda^{s+1} = c\lambda^s$.

**Convergence.** Our algorithm is a natural extension of soft-impute (25), which is a special case of the proximal gradient algorithm for nuclear norm minimization, as demonstrated by (48), and the convergence of the algorithm is guaranteed.

Here we show that the solution of our imputation algorithm converges asymptotically to a minimizer of the objective $L_\lambda(\hat{M})$ in an elegant argument. We show that each step of our imputation algorithm is minimizing a surrogate $Q_\lambda(\hat{M}|\hat{M}^{\text{old}}) = \|A^\Omega + P_\Omega^\perp \mathcal{U}^{-1}\hat{M}^{\text{old}} - \mathcal{U}^{-1}\hat{M}\|^2 + \sum_{k=1}^N \lambda_k \|\hat{M}(k)\|_*$.

**Theorem 2** *The imputation algorithm produces a sequence of iterates $\hat{M}_\lambda^t$ as the minimizer of the successive optimization objective*

$$\hat{M}_\lambda^{t+1} = \text{argmin} \quad Q_\lambda(\hat{M}|\hat{M}_\lambda^t).$$

*The sequence of iterates that converges to the minimizer $\hat{M}^*$ of $L_\lambda(\hat{M})$.*

We put the proof of the convergence theorem in the appendix. The main idea of the proof is to show that

- $Q_\lambda$ decreases after every iteration.
- $\hat{M}_\lambda^t$ is a Cauchy sequence.
- The limit point is a stationary point of $L_\lambda$

**Computational complexity.** Now we analyze the computational complexity of the imputation algorithm. The cost of the graph Fourier transform on matrix network is $O(mnN^2)$. When the graph is a periodic lattice, using fast Fourier transform(FFT), it is reduced to $O(mnN \log N)$. The cost of SVD is $O(\min(mn^2, m^2n)N)$ for computing singular value soft-threshold. Replacing SVD with alternating ridge regression reduces the complexity to $O(r^2nN)$. Therefore, the cost of each iteration is the sum of the cost of both parts, and the total cost would be that times total iteration steps.

## 7. Experimental results

**Numerical verification of the exact recovery** To focus on the essential difficulty of the problem, we study the noiseless, node sampling setting: In each imputation experiment, we first generate a stack of low-rank matrices in the spectral space, $\tilde{A}(k) = X^0(k)^T Y^0(k)$ for i.i.d Gaussian random matrix $X^0(k), Y^0(k) \in \mathbf{R}^{r \times n}$. We also generate a random graph $G$. Then we compute the matrix network $A$ by the inverse graph Fourier transform and obtain our observation by node undersampling of $A$. Then we send the observed matrices and the graph $G$ to the imputation algorithm to get the solution $\hat{M}$. We measure the relative mean square error (rMSE) $\|\hat{M} - \hat{A}\| / \|\hat{A}\|$.

We set $(n, N) = (50, 100)$ for all our experiments, and vary the undersampling ratio $p$ and rank $r$. For each set of parameters $(p, r)$, we repeat the experiment multiple times and compute the success rate of exact recovery. In figure 4 on the upper panel we show the rMSE when the graphs are one-dimensional chains of length $N$. When $r/n$ is large and $p$ is small, the rMSE is approximately equal to the undersampling ratio $p$, which means the optimization

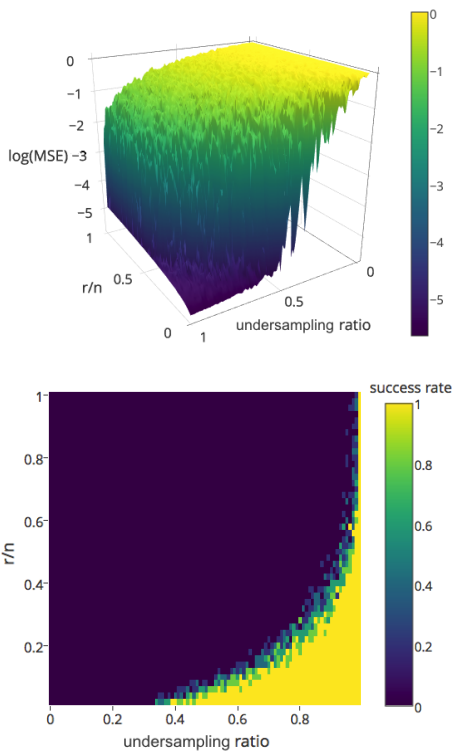

|  | $N_{\text{obs}} = 0.2N$ | $N_{\text{obs}} = 0.4N$ | $N_{\text{obs}} = 0.9N$ |
|  | $p = 1$ | $p = 0.5$ | $p = 2/9$ |
| --- | --- | --- | --- |
| $\sigma = 0$ | 0.116/0.000 | 0.100/0.038 | 0.088/0.061 |
| $\sigma = 0.1$ | 0.363/0.495 | 0.348/0.411 | 0.339/0.365 |

Table 1: Average MSE of missing/(partially) observed matrices.

Figure 4: Upper: the rMSE for different combination of undersampling ratio $p$ and rank $r/n$. We observe that the transition between successful recovery (rMSE $\approx 10^{-5}$) and failure (rMSE $\approx p$) is very sharp. Lower: the phase transition graph with varying undersampling ratio and rank. We repeat the experiment multiple times for each parameter combination, and plot the success recovery rate (rMSE $<$ 0.001)

failed to recover the ground truth matrices. On the opposite side, when $r/n$ is small and $p$ is large, the rMSE is very small, indicating we have successfully recovered the missing matrices. The transition between the two regions is very sharp. We also show the success rate on the lower panel of figure 4, which demonstrates a phase transition.

**Feature matrices on Facebook network** We take the ego networks from the SNAP Facebook dataset (33). The combined network forms a connected graph with 4039 nodes and 88234 edges. All the edges have equal weights. The feature matrices on each of the nodes were generated by randomly generating $X(k), Y(k) \in \mathbf{C}^{1 \times 50}$ in the spectral domain, and doing the inverse graph Fourier transform to get $A = \mathcal{U}^{-1}(X(k)Y(k))$. The observation is generated by sampling $N_{\text{obs}}$ matrices at sampling rate $p$, and adding i.i.d. Gaussian noise with mean 0 and variance $\sigma^2/50$ to all observed entries. Here $N_{\text{obs}} < N = 4039$ and the other matrices are completely unobserved.

We run our iterative imputation algorithm to recover $A$ from this observation with varying parameters $N_{\text{obs}}, p,$ and $\sigma$, and calculate the MSE between our estimation and the ground truth. The results are summarized in Table 1. When there is no additive noise, we can recover all the matrices very well even with only 20% of entries observed across the matrix network. It works well both when doing node undersampling and more uniform undersampling. When there is additive noise, the MSE between reconstruction and the ground truth will grow proportionally.

**MRI completion** We use a cardiac MRI scan dataset for the completion task. The stack of MRI images scans through a human torso. The frames are corrupted, several frames are missing, and the other frames are sampled i.i.d. from a Bernoulli distribution with $p = 0.2$. Our completion result is demonstrated in figure 1 at first page as the motivating example. In the 88 frames there are 2 frames missing, and we only sampled 20% of the rest of frames i.i.d. from a Bernoulli distribution with $p = 0.2$. We compare with the baseline method where we solve a tensor completion problem using nuclear norm minimization. Relative MSE for all frames are plotted in figure 5. The baseline method failed at missed frames and significantly under-performed the convolutional imputation method.

**SPECT completion** We imputed a cardiac SPECT scan dataset. The SPECT scan captures the periodic movement of a heart, and we have a temporal sequence at a fixed spatial slice. The sequence has 36 frames, capturing 4 periods of heart beats. 4 consecutive frames out of the 36 frames are missing and the other frames are sampled i.i.d. from a Bernoulli distribution with $p = 0.2$. We try to recover the whole image stack from the observations and compare our method with two baseline methods. The first baseline method assumes each individual frame is low-rank and minimizes the sum of nuclear norms. The second baseline method adds the graph regularizer from (41; 23; 44), in addition to the low-rank assumption on each frame. Minimizing the sum of nuclear norm fails to recover completely missing frames. Our algorithm performs better than tensor completion with graph regularizer on the SPECT scan, since in spectral domain we can use the periodicity to help aggregate information, while using graph regularizer only propagates information between neighbors. This is demon-strated in figure 6. The first row shows the ground truth, and the second row overlays the ground truth (in red channel) with the completion result using our convolutional imputation algorithm (in green channel). The third row overlays the ground truth with the completion result using tensor completion with graph regularizer. The completion result with our algorithm matches the ground truth very well, while the completion result with tensor completion using graph regularizer is biased towards the average of neighboring frames, showing red and green rings on the edges. A quantitative comparison on the SPECT scan completion is given in figure 7. Our imputation algorithm's relative MSE between reconstruction and the ground truth is significantly smaller than the baselines'. It is worth pointing out that our method's recovery performance at the missing frames are comparable to that at the partially observed frames, while the first baseline completely fails at the missing frames and the second baseline performs significantly worse.

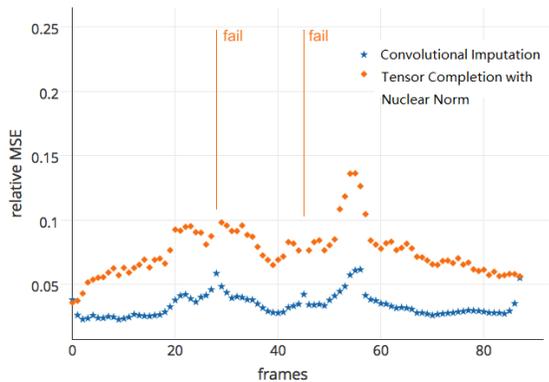

Figure 5: Comparison of relative MSE for all frames of MRI, the baseline joint matrix completion method failed at missed frames and significantly underperformed the convolutional imputation method.

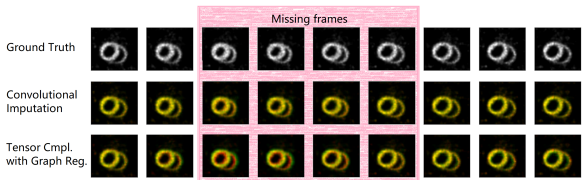

Figure 6: Visualization of the first 9 frames of the SPECT sequence. The frames in pink shadow are missing in the observation.

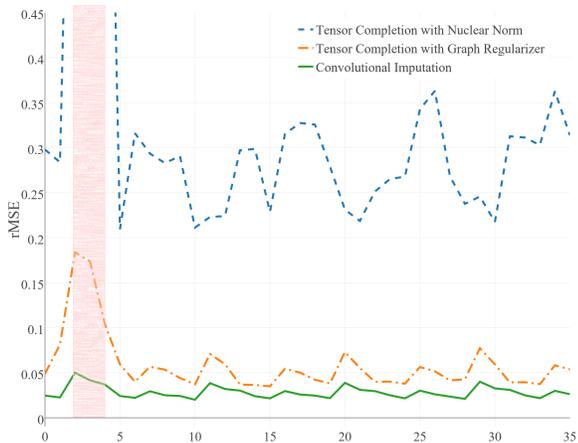

Figure 7: Comparison of relative MSE for all frames of SPECT, missed frames are indexed 3 to 5.

## 8. Discussion

In practice, when you are given a tensor or a stack of matrices, there are two ways to formulate it into a matrix network.

One is to use the knowledge of physical or geometrical relation to naturally determine the graph. The graph of the matrix network is given in the facebook network and the graph is naturally constructed as a 1-d equal-weighted chain in the MRI and SPECT datasets, based on the nature of the datasets.

The other is to construct the graph using an explicit constructive methods. Finding a graph with good graph Fourier transform relies on problem structure and domain knowledge. One suggested universal way is to construct a lattice or a $d-$regular graph, then assign the weight on each edge as some distance metric of two matrices, for example, the distance metric could be computed using Gaussian kernels. We suggest that the coherence $\mu$ we defined before could be used as a criterion to measure how good the graph Fourier transform is. From the bound on $\mu$ by the coherence of the graph Fourier transform and the maximum coherence over all spectral matrices, we know that we want to search for graph with low coherence. This leads to interesting dictionary learning problem where we want to learn a unitary dictionary as the graph Fourier transform.

To conclude, treating a series of matrices with relations as a matrix network is a useful modeling framework since a matrix network has operations like the graph Fourier transform and convolution. This framework allows us to complete the matrices when some of them are completely unobserved, using the spectral low-rank structural assumption. We provided an exact recovery guarantee and discovered a new phase transition phenomenon for the completion algorithm.

## References

[1] Rie Kubota Ando and Tong Zhang. Learning on graph with laplacian regularization. *Advances in neural information processing systems*, 19:25, 2007.

[2] Konstantin Avrachenkov, Pavel Chebotarev, and Alexey Mishenin. Semi-supervised learning with regularized lapla-

**Convolutional Imputation of Matrix Networks**cian. *Optimization Methods and Software*, 32(2):222–236, 2017.

[3] Mohsen Bayati, Marc Lelarge, and Andrea Montanari. Universality in polytope phase transitions and iterative algorithms. *Information Theory Proceedings (ISIT), 2012 IEEE International Symposium on*, pages 1643–1647, 2012.

[4] Mikhail Belkin and Partha Niyogi. Laplacian eigenmaps and spectral techniques for embedding and clustering. In *Advances in neural information processing systems*, pages 585–591, 2002.

[5] Shimon Brooks and Elon Lindenstrauss. Non-localization of eigenfunctions on large regular graphs. *Israel Journal of Mathematics*, 193(1):1–14, 2013.

[6] Joan Bruna, Wojciech Zaremba, Arthur Szlam, and Yann LeCun. Spectral networks and locally connected networks on graphs. *arXiv preprint arXiv:1312.6203*, 2013.

[7] Jian-Feng Cai, Emmanuel Candès, and Zuowei Shen. A singular value thresholding algorithm for matrix completion. *SIAM Journal on Optimization*, 20(4):1956–1982, 2010.

[8] Emmanuel Candès and Yaniv Plan. Matrix completion with noise. *Proceedings of the IEEE*, 98(6):925–936, 2010.

[9] Emmanuel Candès and Benjamin Recht. Exact matrix completion via convex optimization. *Communications of the ACM*, 55(6):111–119, 2012.

[10] Emmanuel Candès and Terence Tao. The power of convex relaxation: Near-optimal matrix completion. *IEEE Transactions on Information Theory*, 56(5):2053–2080, 2010.

[11] Fan RK Chung. *Spectral graph theory*. Number 92. American Mathematical Soc., 1997.

[12] Ronald R Coifman, Stephane Lafon, Ann B Lee, Mauro Maggioni, Boaz Nadler, Frederick Warner, and Steven W Zucker. Geometric diffusions as a tool for harmonic analysis and structure definition of data: Diffusion maps. *Proceedings of the National Academy of Sciences of the United States of America*, 102(21):7426–7431, 2005.

[13] Ronald R Coifman and Mauro Maggioni. Diffusion wavelets. *Applied and Computational Harmonic Analysis*, 21(1):53–94, 2006.

[14] Yael Dekel, James R Lee, and Nathan Linial. Eigenvectors of random graphs: Nodal domains. *Random Structures & Algorithms*, 39(1):39–58, 2011.

[15] David Donoho, Matan Gavish, et al. Minimax risk of matrix denoising by singular value thresholding. *The Annals of Statistics*, 42(6):2413–2440, 2014.

[16] David Donoho and Jared Tanner. Counting faces of randomly projected polytopes when the projection radically lowers dimension. *Journal of the American Mathematical Society*, 22(1):1–53, 2009.

[17] Ioana Dumitriu, Soumik Pal, et al. Sparse regular random graphs: spectral density and eigenvectors. *The Annals of Probability*, 40(5):2197–2235, 2012.

[18] Ahmed El Alaoui, Xiang Cheng, Aaditya Ramdas, Martin J Wainwright, and Michael I Jordan. Asymptotic behavior of lp-based laplacian regularization in semi-supervised learning. *29th Annual Conference on Learning Theory*, pages 879–906, 2016.

[19] Marko Filipović and Ante Jukić. Tucker factorization with missing data with application to low-rank tensor completion. *Multidimensional systems and signal processing*, 26(3):677–692, 2015.

[20] Jacob Fox, Tim Roughgarden, C Seshadhri, Fan Wei, and Nicole Wein. Finding cliques in social networks: A new distribution-free model. *arXiv preprint arXiv:1804.07431*, 2018.

[21] Silvia Gandy, Benjamin Recht, and Isao Yamada. Tensor completion and low-n-rank tensor recovery via convex optimization. *Inverse Problems*, 27(2):025010, 2011.

[22] Matan Gavish and David L Donoho. The optimal hard threshold for singular values is $4/\sqrt{(3)}$. *IEEE Transactions on Information Theory*, 60(8):5040–5053, 2014.

[23] Pere Giménez-Febrer and Alba Pages-Zamora. Matrix completion of noisy graph signals via proximal gradient minimization.

[24] David Gross. Recovering low-rank matrices from few coefficients in any basis. *IEEE Transactions on Information Theory*, 57(3):1548–1566, 2011.

[25] T Hastie and R Mazumder. softimpute: Matrix completion via iterative soft-thresholded svd. *R package version*, 1, 2015.

[26] Trevor Hastie, Rahul Mazumder, Jason D Lee, and Reza Zadeh. Matrix completion and low-rank svd via fast alternating least squares. *J. Mach. Learn. Res*, 16(1):3367–3402, 2015.

[27] Mikael Henaff, Joan Bruna, and Yann LeCun. Deep convolutional networks on graph-structured data. *arXiv preprint arXiv:1506.05163*, 2015.

[28] Wei Hu, Gene Cheung, Antonio Ortega, and Oscar C Au. Multiresolution graph fourier transform for compression of piecewise smooth images. *IEEE Transactions on Image Processing*, 24(1):419–433, 2015.

[29] M.O. Jackson. *Social and Economic Networks*. Princeton University Press. Princeton University Press, 2008.

[30] Eric Kernfeld, Misha Kilmer, and Shuchin Aeron. Tensor–tensor products with invertible linear transforms. *Linear Algebra and its Applications*, 485:545–570, 2015.

[31] Raghunandan H Keshavan, Andrea Montanari, and Sewoong Oh. Matrix completion from a few entries. *IEEE Transactions on Information Theory*, 56(6):2980–2998, 2010.

[32] Raghunandan H Keshavan, Andrea Montanari, and Sewoong Oh. Matrix completion from noisy entries. *Journal of Machine Learning Research*, 11(Jul):2057–2078, 2010.

[33] Jure Leskovec and Julian J Mcauley. Learning to discover social circles in ego networks. *Advances in neural information processing systems*, pages 539–547, 2012.

[34] Ji Liu, Przemyslaw Musialski, Peter Wonka, and Jieping Ye. Tensor completion for estimating missing values in visual data. *IEEE Transactions on Pattern Analysis and Machine Intelligence*, 35(1):208–220, 2013.

[35] Xiao-Yang Liu, Shuchin Aeron, Vaneet Aggarwal, Xiaodong Wang, and Min-You Wu. Adaptive sampling of rf fingerprints for fine-grained indoor localization. *IEEE Transactions on Mobile Computing*, 15(10):2411–2423, 2016.

[36] Rahul Mazumder, Trevor Hastie, and Robert Tibshirani. Spectral regularization algorithms for learning large incomplete matrices. *Journal of machine learning research*, 11(Aug):2287–2322, 2010.

[37] Hatef Monajemi and David Donoho. Sparsity/undersampling tradeoffs in anisotropic undersampling, with applications in mr imaging/spectroscopy. *arXiv preprint arXiv:1702.03062*, 2017.

[38] Boaz Nadler, Nathan Srebro, and Xueyuan Zhou. Semi-supervised learning with the graph laplacian: The limit of infinite unlabelled data. *Advances in neural information processing systems*, 21, 2009.

[39] Samet Oymak and Babak Hassibi. The proportional mean decomposition: a bridge between the gaussian and bernoulli ensembles. *Acoustics, Speech and Signal Processing (ICASSP), 2015 IEEE International Conference on*, pages 3322–3326, 2015.

[40] Nathanael Perraudin, Benjamin Ricaud, David Shuman, and Pierre Vandergheynst. Global and local uncertainty principles for signals on graphs. *arXiv preprint arXiv:1603.03030*, 2016.

[41] Nikhil Rao, Hsiang-Fu Yu, Pradeep K Ravikumar, and Inderjit S Dhillon. Collaborative filtering with graph information: Consistency and scalable methods. *Advances in neural information processing systems*, pages 2107–2115, 2015.

[42] Naoki Saito and Ernest Woei. On the phase transition phenomenon of graph laplacian eigenfunctions on trees (recent development and scientific applications in wavelet analysis). 2011.

[43] Aliaksei Sandryhaila and José MF Moura. Discrete signal processing on graphs: Graph fourier transform. *ICASSP*, pages 6167–6170, 2013.

[44] Nauman Shahid, Nathanael Perraudin, Gilles Puy, and Pierre Vandergheynst. Compressive pca for low-rank matrices on graphs. *IEEE transactions on Signal and Information Processing over Networks*, 2016.

[45] David I Shuman, Sunil K Narang, Pascal Frossard, Antonio Ortega, and Pierre Vandergheynst. The emerging field of signal processing on graphs: Extending high-dimensional data analysis to networks and other irregular domains. *IEEE Signal Processing Magazine*, 30(3):83–98, 2013.

[46] David I Shuman, Benjamin Ricaud, and Pierre Vandergheynst. Vertex-frequency analysis on graphs. *Applied and Computational Harmonic Analysis*, 40(2):260 – 291, 2016.

[47] Linh V Tran, Van H Vu, and Ke Wang. Sparse random graphs: Eigenvalues and eigenvectors. *Random Structures & Algorithms*, 42(1):110–134, 2013.

[48] Quanming Yao and James T Kwok. Accelerated inexact soft-impute for fast large-scale matrix completion. *Twenty-Fourth International Joint Conference on Artificial Intelligence*, 2015.

[49] Xiaoqin Zhang, Di Wang, Zhengyuan Zhou, and Yi Ma. Simultaneous rectification and alignment via robust recovery of low-rank tensors. In *Advances in Neural Information Processing Systems*, pages 1637–1645, 2013.

[50] Xiaoqin Zhang, Zhengyuan Zhou, Di Wang, and Yi Ma. Hybrid singular value thresholding for tensor completion. In *Twenty-Eighth AAAI Conference on Artificial Intelligence*, 2014.

[51] Zemin Zhang and Shuchin Aeron. Exact tensor completion using t-svd. *IEEE Transactions on Signal Processing*, 65(6):1511–1526, 2016.

[52] Zhengyuan Zhou, Nicholas Bambos, and Peter Glynn. Dynamics on linear influence network games under stochastic environments. In *International Conference on Decision and Game Theory for Security*, pages 114–126. Springer, 2016.

[53] Zhengyuan Zhou, Nicholas Bambos, and Peter Glynn. Deterministic and stochastic wireless networks games: Equilibrium, dynamics and price of anarchy. *Operations Research*, 2018.

[54] Zhengyuan Zhou, Benjamin Yolken, Reiko Ann Miura-Ko, and Nicholas Bambos. A game-theoretical formulation of influence networks. In *American Control Conference (ACC), 2016*, pages 3802–3807. IEEE, 2016.

[55] Xiaofan Zhu and Michael Rabbat. Approximating signals supported on graphs. *Acoustics, Speech and Signal Processing (ICASSP), 2012 IEEE International Conference on*, pages 3921–3924, 2012.

# Appendix: Convolutional Imputation of Matrix Networks

Qingyun Sun [*1]  Mengyuan Yan [*2]  David Donoho [3]  Stephen Boyd [2]

**Exact recovery guarantee**

**Theorem 1.** *We assume that $A$ is a matrix network on a graph $G$, and its graph Fourier transform $\hat{A}(k)$ are a sequence of matrices, each of them is at most rank $r$, and $\hat{A}$ satisfy the incoherence condition with coherence $\mu$. And we observe a matrix network $A^\Omega$ on the graph $G$, for a subset of node in $\Omega$ random sampled from the network, node $i$ on the network is sampled with probability $p_i$, we define the average sampling rate $p = \frac{1}{N}\sum_{i=1}^N p_i = |\Omega|/(Nn^2)$, and define $\mathcal{R} = \frac{1}{p}P_\Omega \mathcal{U}^*$.*

*Then we prove that for any sampling probability distribution $\{p_i\}$, as long as the average sampling rate $p > C\mu \frac{r}{n} \log^2(Nn)$ for some constants $C$, the solution to the optimization problem*

$$\begin{aligned} \underset{\hat{M}}{\text{minimize}} \quad & \|\hat{M}\|_{*,1}, \\ \text{subject to} \quad & A^\Omega = \mathcal{R}\hat{M} \end{aligned}$$

*is unique and is exactly $\hat{A}$ with probability $1 - (Nn)^{-\gamma}$, where $\gamma = \frac{\log(Nn)}{16}$.*

*Proof.* We define a inner product: $\langle \hat{M}_1, \hat{M}_2 \rangle = \sum_k \langle \hat{M}_1(k), \hat{M}_2(k) \rangle$. Then we have the following two inequalities

$$\|\hat{M}(k)\|_* = \mathbf{Tr}(\text{sgn}(\hat{M}(k))\hat{M}(k)) = \langle \text{sgn}(\hat{M}(k)), \hat{M}(k)\rangle.$$

Therefore,

$$\|\hat{M}\|_{*,1} = \langle \text{sgn}(\hat{M}), \hat{M}\rangle.$$

Here $\text{sgn}(\hat{M}) = V_1 V_2^*$ is the sign matrix of the singular values of $\hat{M}$ under the singular vector basis.

We consider $\Delta = \hat{M} - \hat{A}$, then either $\mathcal{R}\Delta \neq 0$, or $\|\hat{A} + \Delta\|_{*,1} > \|\hat{A}\|_{*,1}$.

*Equal contribution  [1]Department of Mathematics, Stanford University, California, USA  [2]Department of Electrical Engineering, Stanford University, California, USA  [3]Department of Statistics, Stanford University, California, USA. Correspondence to: Qingyun Sun <qysun@stanford.edu>.

First we define a decomposition $\Delta = \Delta_T + \Delta_T^\perp = P_T\Delta + P_{T^\perp}\Delta$.

For $\mathcal{R}\Delta = 0$, we compute

$$\begin{aligned} & \|\hat{A} + \Delta\|_{*,1} \\ \geq\ & \|P_1(\hat{A}+\Delta)P_2\|_{*,1} + \|P_1^\perp(\hat{A}+\Delta)P_2^\perp\|_{*,1} \\ =\ & \|\hat{A} + P_1\Delta P_2\|_{*,1} + \|\Delta_T^\perp\|_{*,1} \\ \geq\ & \langle \text{sgn}(\hat{A}), \hat{A} + P_1\Delta P_2\rangle + \langle \text{sgn}(\Delta_T^\perp), \Delta_T^\perp\rangle \\ =\ & \|\hat{A}\|_{*,1} + \langle \text{sgn}(\hat{A}), P_1\Delta P_2\rangle + \langle \text{sgn}(\Delta_T^\perp), \Delta_T^\perp\rangle \\ =\ & \|\hat{A}\|_{*,1} + \langle \text{sgn}(\hat{A}) + (\Delta_T^\perp), \Delta\rangle. \end{aligned}$$

Now we want to estimate $\langle (\hat{A}) + (\Delta_T^\perp), \Delta\rangle$. We make two assumptions, which we will prove later.

First, we assume that for all $\Delta \in \text{range}(\mathcal{R})^\perp$, with probability $1 - (Nn)^{-\gamma}$,

$$\|\Delta_T\|_2 < 2nN\|\Delta_T^\perp\|_2.$$

Second, we want to construct a dual certificate $K \in \text{range}(\mathcal{R})$, such that for $k = 3 + \frac{1}{2}\log_2(r) + \log_2(n) + \log_2(N)$, with probability $1 - (Nn)^{-\gamma}$,

$$\begin{aligned} \|P_T(K) - \mathbf{sgn}(\hat{A})\|_2 &\leq\ (\tfrac{1}{2})^k\sqrt{r}, \\ \|P_{T^\perp}(K)\| &\leq\ \tfrac{1}{2}. \end{aligned}$$

Then

$$\begin{aligned} & \langle \text{sgn}(\hat{A}) + (\Delta_T^\perp), \Delta\rangle \\ =\ & \langle \text{sgn}(\hat{A}) + (\Delta_T^\perp) - K, \Delta\rangle \\ =\ & \langle \text{sgn}(\hat{A}) - K, \Delta_T\rangle + \langle (\Delta_T^\perp) - K, \Delta_T^\perp\rangle \\ \geq\ & \tfrac{1}{2}\|\Delta_T^\perp\|_2 - (\tfrac{1}{2})^k\sqrt{r}\|\Delta_T\|_2 \\ \geq\ & \tfrac{1}{4}\|\Delta_T^\perp\|_2. \end{aligned}$$

When $\hat{M}$ is a minimizer, we must have $\Delta_T^\perp = 0$, otherwise $\|\hat{A}+\Delta\|_{*,1} < \|\hat{A}\|_{*,1}$. By assumption, $\|\Delta_T\|_2 < n^2\|\Delta_T^\perp\|_2$., $\Delta_T = 0$, then $\Delta = 0$. Therefore, under the two assumption, $\hat{M}$ is the unique mininizer, and $\hat{M} = \hat{A}$.

Now we prove the above assumption and construct dual certificate.

First, we show that if

$$\|\Delta_T\|_2 \geq (2nN)\|\Delta_T^\perp\|_2,$$

then $\|\mathcal{R}\Delta_T\|_2 > \|\mathcal{R}\Delta_T^\perp\|_2$,

$$\begin{aligned}\|\mathcal{R}\Delta\|_2 &= \|\mathcal{R}\Delta_T + \mathcal{R}\Delta_T^\perp\|_2 \\ &\geq \|\mathcal{R}\Delta_T\|_2 - \|\mathcal{R}\Delta_T^\perp\|_2 \\ &> 0.\end{aligned}$$

We have a lower bound on $\|\mathcal{R}\Delta_T\|_2$ and upper bound on $\|\mathcal{R}\Delta_T^\perp\|_2$.

$$\|\mathcal{R}\Delta_T^\perp\|_2^2 \leq \|\mathcal{R}\|^2 \|\Delta_T^\perp\|_2^2.$$

Here $\|\mathcal{R}\|$ is the operator norm of $\mathcal{R}$.

$$\begin{aligned}\|\mathcal{R}\Delta_T\|_2^2 &= \langle \mathcal{R}\Delta_T, \mathcal{R}\Delta_T\rangle \\ &\geq \|\mathcal{R}\|^2/(nN)^2 (1 - \|P_T - P_T\mathcal{R}P_T\|)\|\Delta_T\|_2^2.\end{aligned}$$

Since $E(P_T\mathcal{R}P_T) = P_T$, we only need to control the deviation, we could use a concentration inequality called operator-Bernstein inequality (1),

$$\mathbf{P}[\|P_T - P_T\mathcal{R}P_T\| > t] \leq \exp(-\frac{npt^2}{4\mu r}).$$

Using the condition that $p = C\mu\frac{r}{n}\log^2(Nn)$, let $t = 1/4$, we have

$$\begin{aligned}\mathbf{P}[\|P_T - P_T\mathcal{R}P_T\| > t] &\leq \exp(-\frac{n\mu\frac{r}{n}\log^2(Nn)}{16\mu r}) \\ &= \exp(-\frac{\log^2(Nn)}{16}) \\ &= (nN)^{-\gamma},\end{aligned}$$

where $\gamma = \frac{\log(Nn)}{16}$. Therefore, with probability $1 - (nN)^{-\gamma}$, the the inequality holds for $t = 1/2$. When the inequality holds, $\|P_T - P_T\mathcal{R}P_T\| < 1/2$, $\mathcal{R}\Delta \neq 0$.

Second, we construct the dual certificate $K$ by the following construction: We decompose $\Omega$ as the union of $k$ subset $\Omega_t$, where each entry is sampled independently so that $E(|\Omega_t| = p_t = 1 - (1-p)^{1/k}$, and define $R_t = \frac{1}{p_t}P_{\Omega_t}\mathcal{U}^*$. Define

$$H_0 = (\hat{A}), K_t = \sum_{j=1}^t R_j H_{j-1}, H_t = (\hat{A}) - P_T K_t.$$

Then the dual certificate is defined as $K = K_k$.

This construction is called golfing scheme, which is invented in (1). Since $p_t = p/k = C\mu\frac{r}{nk}\log^2(Nn)$, we can assume $\|P_T - P_T\mathcal{R}_j P_T\| < 1/2$, which is true with probability $1 - \exp(\frac{Cnpt^2}{\mu kr})$.

$$\|H_t\|_2 \leq \|P_T - P_T\mathcal{R}P_T\|\|H_{t-1}\|_2 \leq \frac{1}{2}\|H_{t-1}\|_2.$$

And

$$\|P_T(K) - (\hat{A})\|_2 = \|H_k\| \leq (\frac{1}{2})^k\|(\hat{A})\| \leq (\frac{1}{2})^k\sqrt{r}.$$

Then

$$\|P_T(K) - (\hat{A})\|_2 \leq (\frac{1}{2})^k\sqrt{r}.$$

Also, $\|P_{T^\perp}(K)\| \leq \sum_{j=1}^k \|P_{T^\perp}R_jH_{j-1}\|$, use the operator-Bernstein inequality for a sequence of $t_j = 1/(4\sqrt{r})$, we have $\|P_{T^\perp}R_jH_{j-1}\| \leq t_i\|H_{j-1}\|_2$, and since $\|H_j\|_2 \leq \sqrt{r}2^{-j}$, then

$$\|P_{T^\perp}(K)\| \leq \sum_{j=1}^k t_i\|H_{j-1}\|_2 \leq \frac{1}{4}\sum_{j=1}^k 2^{-(j-1)} < 1/2.$$

Therefore, $K$ is the dual certificate, the whole proof is done.

$\square$

**Imputation algorithm convergence** Now we show that the solution of our imputation algorithm converges asymptotically to a minimizer of the previously defined objective $L_\lambda(\hat{M})$.

Each step of our imputation algorithm is minimizing a surrogate $Q_\lambda(\hat{M}|\hat{M}^{\text{old}})$ of the above objective function as

$$\|A^\Omega + P_\Omega^\perp \mathcal{U}^{-1}\hat{M}^{\text{old}} - \mathcal{U}^{-1}\hat{M}\|^2 + \sum_{k=1}^N \lambda_k\|\hat{M}(k)\|_*.$$

The resulting minimizer forms a sequence $\hat{M}_\lambda^t$ with starting point $\hat{M}_\lambda^0$

$$\hat{M}_\lambda^{t+1} = \operatorname{argmin} \quad Q_\lambda(\hat{M}|\hat{M}_\lambda^t).$$

**Theorem 2.** *The imputation algorithm produces a sequence of iterates $\hat{M}_\lambda^t$ that converges to the minimizer of $L_\lambda(\hat{M})$.*

The main idea of the proof is to show that $Q_\lambda$ decreases after every iteration and $\hat{M}_\lambda^t$ is a Cauchy sequence, and the limit point is a stationary point of $L_\lambda$.

*Proof.* For each iteration in our algorithm, we are solving for a surrogate of the objective function as

$$Q_\lambda(\hat{M}|\hat{M}^{\text{old}}) = \|A^\Omega + P_\Omega^\perp \mathcal{U}^{-1}\hat{M}^{\text{old}} - \mathcal{U}^{-1}\hat{M}\|^2 + \sum_{k=1}^N \lambda_k\|\hat{M}(k)\|_*.$$

And the sequence $\hat{M}_\lambda^t$ with any starting point $\hat{M}_\lambda^0$ is given by

$$\hat{M}_\lambda^{t+1} = \operatorname{argmin} \quad Q_\lambda(\hat{M}|\hat{M}_\lambda^t).$$

The sequence satisfies

$$L_\lambda(\hat{M}_\lambda^{t+1}) \leq Q_\lambda(\hat{M}_\lambda^{t+1}|\hat{M}_\lambda^t) \leq L_\lambda(\hat{M}_\lambda^t).$$

Because

$$Q_\lambda(\hat{M}_\lambda^{t+1}|\hat{M}_\lambda^{t+1}) = L_\lambda(\hat{M}_\lambda^{t+1})$$

and

$$Q_\lambda(\hat{M}|\hat{M}^{\text{old}})$$
$$= \|P_\Omega(A) + P_\Omega^\perp \mathcal{U}^{-1}\hat{M}^{\text{old}} - \mathcal{U}^{-1}\hat{M}\|^2 + \sum_{k=1}^N \lambda_k \|\hat{M}(k)\|_*$$
$$\geq \|P_\Omega(A) + P_\Omega^\perp \mathcal{U}^{-1}\hat{M} - \mathcal{U}^{-1}\hat{M}\|^2 + \sum_{k=1}^N \lambda_k \|\hat{M}(k)\|_*$$
$$= Q_\lambda(\hat{M}|\hat{M})$$

Below we prove the following successive differences are monotonically decreasing

$$\|\hat{M}_\lambda^{t+1} - \hat{M}_\lambda^t\|^2 \leq \|\hat{M}_\lambda^t - \hat{M}_\lambda^{t-1}\|^2.$$

and the difference sequence converges to zero,

$$\hat{M}_\lambda^{t+1} - \hat{M}_\lambda^t \to 0.$$

The successive differences are monotonically decreasing because the soft threshold operator is a contraction in $L_2$ norm (2). And when there are positive singular values smaller than the threshold, the successive differences will strictly decrease until the algorithm converges.

Then $\hat{M}_\lambda^t$ is a Cauchy sequence, therefore we have a set of limit points. Also by monotonic convergence theorem, since $\hat{M}_\lambda^{t+1} - \hat{M}_\lambda^t$ converges to zero monotonically, the Cauchy sequence $\hat{M}_\lambda^t$ has an unique limit $\hat{M}_\lambda^\infty$. Moreover, we can verify that $\hat{M}_\lambda^\infty$ is a solution to the fixed point equation $\nabla L_\lambda = 0$, and a stationary point of $L_\lambda(\hat{M}_\lambda)$. Since $L_\lambda(\hat{M}_\lambda)$ is convex, each stationary point is a minimizer. Therefore, the convergence is proved. □

## References

[1] David Gross. Recovering low-rank matrices from few coefficients in any basis. *IEEE Transactions on Information Theory*, 57(3):1548–1566, 2011.

[2] Rahul Mazumder, Trevor Hastie, and Robert Tibshirani. Spectral regularization algorithms for learning large incomplete matrices. *Journal of machine learning research*, 11(Aug):2287–2322, 2010.